\documentclass[a4paper, 10pt, conference]{IEEEtran}
\IEEEoverridecommandlockouts
\usepackage{cite}
\usepackage{amsmath,amssymb,amsfonts}
\usepackage{algorithmic}
\usepackage{graphicx}
\usepackage{textcomp}
\usepackage{xcolor}
\usepackage{longtable}

\def\BibTeX{{\rm B\kern-.05em{\sc i\kern-.025em b}\kern-.08em
    T\kern-.1667em\lower.7ex\hbox{E}\kern-.125emX}}

\renewcommand\IEEEkeywordsname{\\ Keywords}

                  \usepackage[english]{babel}
                  \usepackage[utf8]{inputenc}
                  \usepackage{fancyhdr}
\begin{document}

\title
              {A Comparative Study of Sentiment Analysis Using NLP and Different Machine Learning Techniques on US Airline Twitter Data
              }

              \author{\IEEEauthorblockN{\textsuperscript{1}Md Taufiqul Haque Khan Tusar,  \textsuperscript{2}Md. Touhidul Islam}
              \IEEEauthorblockA{\textit{Department of Computer Science and Engineering} \\
              \textit{City University,}
              Dhaka-1216, Bangladesh \\
              \textsuperscript{1}taufiq.deeplearning@gmail.com, \textsuperscript{2}touhid.cse@cityuniversity.edu.bd}
              }

\maketitle

\begin{abstract}
Today’s business ecosystem has become very competitive. Customer satisfaction has become a major focus for business growth. Business organizations are spending a lot of money and human resources on various strategies to understand and fulfill their customer’s needs. But, because of defective manual analysis on multifarious needs of customers, many organizations are failing to achieve customer satisfaction. As a result, they are losing customer's loyalty and spending extra money on marketing. We can solve the problems by implementing Sentiment Analysis. It is a combined technique of Natural Language Processing (NLP) and Machine Learning (ML). Sentiment Analysis is broadly used to extract insights from wider public opinion behind certain topics, products, and services. We can do it from any online available data. 
In this paper, we have introduced two NLP techniques (Bag-of-Words and TF-IDF) and various ML classification algorithms (Support Vector Machine, Logistic Regression, Multinomial Naive Bayes, Random Forest) to find an effective approach for Sentiment Analysis on a large, imbalanced, and multi-classed dataset. Our best approaches provide 77\% accuracy using Support Vector Machine and Logistic Regression with Bag-of-Words technique. 
\end{abstract}

 \begin{IEEEkeywords}
Sentiment Analysis, Machine Learning, SVM, Logistic Regression, Airline, Twitter
\end{IEEEkeywords}

\section{Introduction}
% 		1. What is the problem?
Customer satisfaction is an assessment of consumer's perception of products, services, and organizations. Many researchers have found that the quality of products or services and customer happiness are the most essential aspects of business performance [1]. To ensure the organization’s competitiveness, businesses must carefully consider what their customers require and want from the products or services they provide. Also, they must well manage their customers by making them satisfied to do business with them [2]. In [3], the author investigated data from 2007 to 2011 of the top 14 U.S. airline's service quality and customer satisfaction. The result reveals that the airline sector has been struggling to provide outstanding services and meet the requirements of diverse consumer groups.\\

% 		2. Why it is important to solve?
Most of the data in social networks or any other platforms are unstructured. Extracting customer's opinions and taking necessary decisions from such data is laborious. Sentiment Analysis is a decisive approach that aids in the detection of people’s opinion. The principal aim of Sentiment Analysis is to classify the polarity of textual data, whether it is positive, negative, or neutral. Sentiment Analysis tools enable decision-makers to track changes in public or customer sentiment regarding entities, activities, products, technologies, and services [4]. A business organization can easily improve its products and services, a political party or social organization can achieve quality work with help of Sentiment Analysis. Through Sentiment Analysis, it’s easier to understand broad public opinion in a short time. \\

% 		3. Why it is difficult to solve?
Most of the data for sentiment analysis are collected from social media platforms and stored in files that are called datasets. But it becomes challenging to analyze sentiment when the datasets are imbalanced, large, multi-classed, etc. \\

%		4. What is our approach? 
In this paper, we have worked with a large, imbalanced, multi-classed, and real-world dataset named \textbf{Twitter US Airline Sentiment} [13]. We have applied NLP techniques to pre-process and vectorize the data. Thereafter classified the polarity of textual data using Machine Learning classification algorithms. Applied algorithms are Support Vector Machine, Multinomial Naive Bayes, Random Forest, and Logistic Regression. NLP techniques are Bag-of-Words, Term Frequency - Inverse Document Frequency. Finally, compared the applied Machine Learning algorithms and NLP techniques to find the best approach.
%		5. How it is different from others? 
\section{Methodology}
There are the steps for our approaches:
\begin{enumerate}
\item Collecting dataset to train and test ML Classifier. 
\item  Pre-processing the dataset for subsequent processing. 
\item Converting textual data into vector form using NLP.
\item Dividing the dataset into training and testing groups. Then train the ML Classifier with training data and predict the polarity of testing data.
\end{enumerate}
Fig.1 depicts the workflow of Sentiment Analysis using NLP and different Machine Learning techniques.
\begin{figure}[hbt!]
\centerline{\includegraphics[height=224pt]{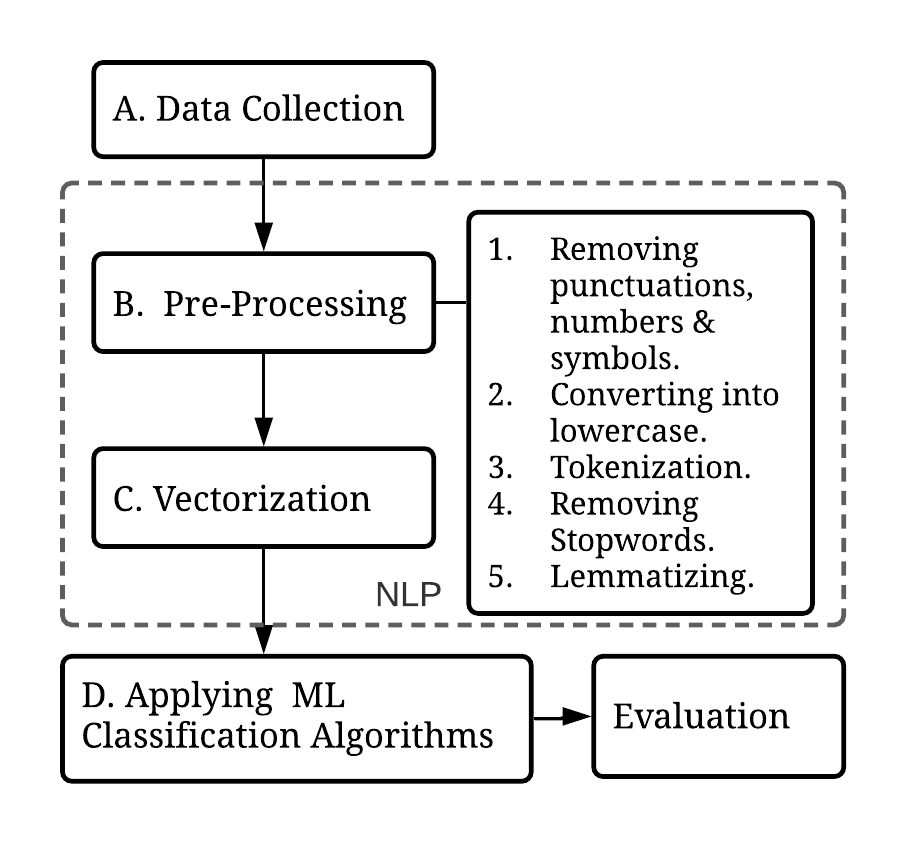}}
\caption{Workflow of Sentiment Analysis using NLP and Machine Learning}
\label{figure}
\end{figure}

\subsection{Data Collection}\label{AA}
The data originally came from CrowdFlower’s Data for Everyone library. Contributors scraped Twitter data of the travelers who traveled through six US airlines in February 2015. They provided the data on Kaggle as a dataset, named \textbf{Twitter US Airline Sentiment} [13] under the \textbf{CC BY-NC-SA 4.0} license. The dataset has around 14640 records and 15 attributes. It contains whether the sentiment of the tweets in this set was positive, neutral, or negative for six US airlines services. Fig.2 shows the frequency of polarity in the dataset.
\begin{figure}[hbt!]
\centerline{\includegraphics[width=1\columnwidth]{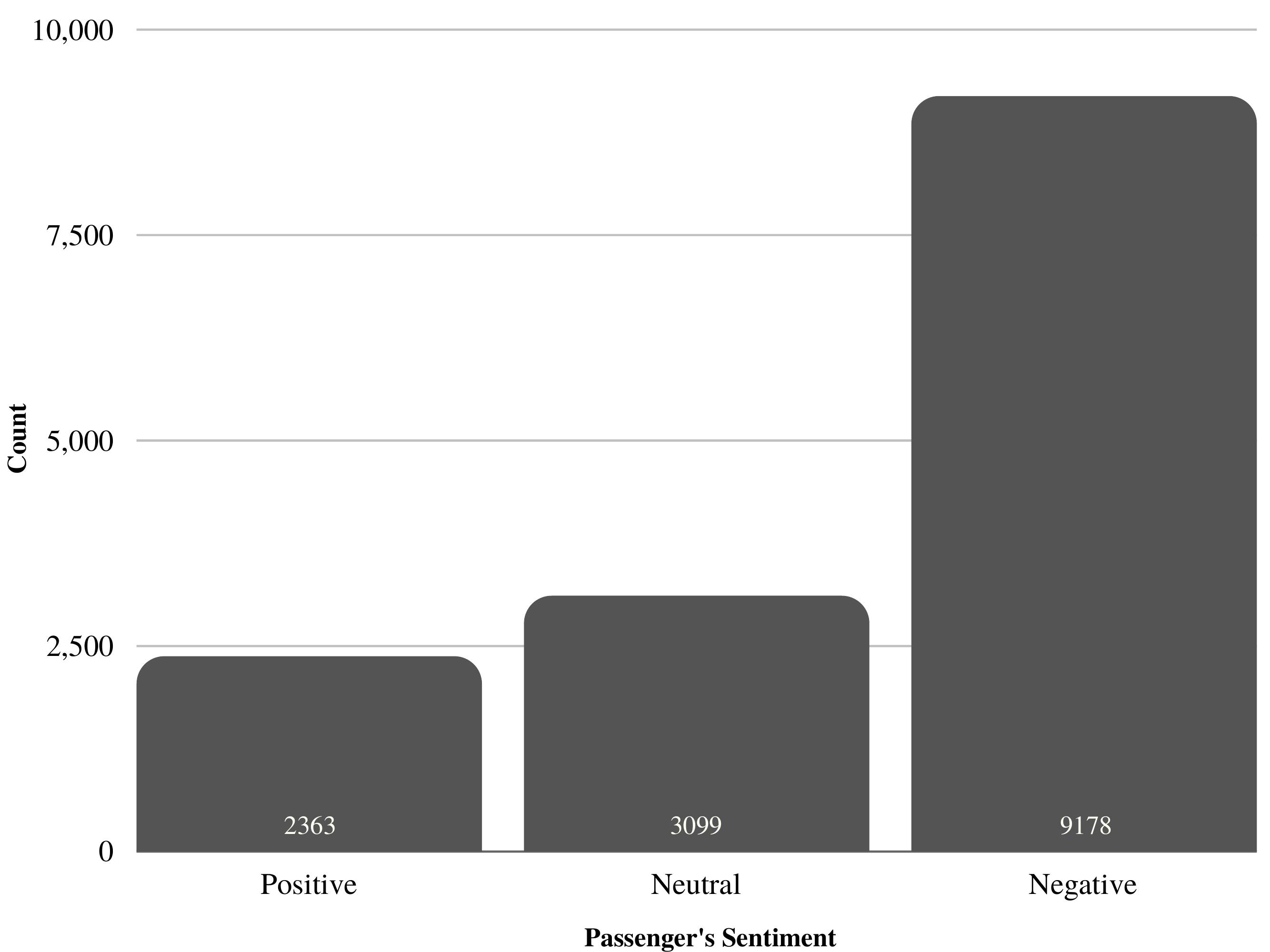}}
\caption{Frequency of Positive, Negative, and Neutral tweets in the dataset}
\label{figure}
\end{figure}

\subsection{Pre-Processing}
A tweet can contain various symbols (!, \#, @, etc), numbers, punctuation, or stop-words. Stop-words mean which words don’t comprise any sentiment. Such as he, she, the, is, that. These are noisy data for Sentiment Analysis. So, we have cleaned the data for further processing by removing punctuation, number, symbol, converting all the characters into lowercase. Then we have divided the tweet into tokens and removed stop-words from the list of tokens. Then converted the tokens into their base form. To convert into the base form, the Lemmatization technique has been used. Then we have stored the cleaned and pre-processed base forms of each tweet in a list called vocabulary. Table I shows the outcome of pre-processing as an example.

\begin{table}[hbt!]
\caption{Pre-processing of noisy data}
\label{tab:my-table}
\begin{tabular}{|c|l|}
\hline
\textbf{Tweet 1}     & \begin{tabular}[c]{@{}l@{}}\#Delicious \#Beef \#Cheese \#Burger \\ @McDonald Testing CheeseBurger and Hamburger\end{tabular}\\ \hline
After Pre-processing & \begin{tabular}[c]{@{}l@{}}{[}delicious, beef, cheese, burger, mcdonald, taste, \\ cheeseburger, hamburger{]}\end{tabular}            \\ \hline
\textbf{Tweet 2}     & \begin{tabular}[c]{@{}l@{}}\#Late Service @McDonald \\ Delicious Hamburger but slow service\end{tabular}                              \\ \hline
After Pre-processing          & {[}late, service, mcdonald, delicious, hamburger, slow{]}                                                                             \\ \hline
\textbf{Vocabulary} &
  \begin{tabular}[c]{@{}l@{}}{[}delicious, beef, cheese, burger, mcdonald, taste,\\ cheeseburger, hamburger, late, service, slow{]}\end{tabular} \\ \hline
\end{tabular}
\end{table} 

\subsection{Vectorization}
The Machine Learning model can not understand the textual data. We have to feed numerical value to the machine learning model. So we should convert the textual data into vector form for subsequent processing. There are two popular Natural Language Processing techniques for Vectorization (i) Bag-of-Words (ii) Term Frequency - Inverse Document Frequency.

\begin{itemize}
\item \textit{Bag-of-Words (BoW): }
The idea behind BoW is to mark the occurrence of the word in each tweet from the vocabulary to convert it into a vector representation. We should use 1s and 0s to mark the appearance of each of these words. Below given an example of Bag-of-Words for tweet 1 and tweet 2 of Table I.

% \usepackage{lscape}

\begin{table}[hbt!]
\caption{Bag of words}
\label{tab:my-table}
\begin{center}
\begin{tabular}{|l|l|l|}
\hline
{\color[HTML]{20124D} \textbf{Tokens}} & {\color[HTML]{20124D} \textbf{Tweet 1}} & {\color[HTML]{20124D} \textbf{Tweet 2}} \\ \hline
{\color[HTML]{20124D} delicious}       & {\color[HTML]{20124D} 1}       & {\color[HTML]{20124D} 1}                \\ \hline
{\color[HTML]{20124D} beef}         & {\color[HTML]{20124D} 1} & {\color[HTML]{20124D} 0} \\ \hline
{\color[HTML]{20124D} cheese}       & {\color[HTML]{20124D} 1} & {\color[HTML]{20124D} 0} \\ \hline
{\color[HTML]{20124D} burger}       & {\color[HTML]{20124D} 1} & {\color[HTML]{20124D} 0} \\ \hline
{\color[HTML]{20124D} mcdonald}     & {\color[HTML]{20124D} 1} & {\color[HTML]{20124D} 1} \\ \hline
{\color[HTML]{20124D} taste}        & {\color[HTML]{20124D} 1} & {\color[HTML]{20124D} 0} \\ \hline
{\color[HTML]{20124D} cheeseburger} & {\color[HTML]{20124D} 1} & {\color[HTML]{20124D} 0} \\ \hline
{\color[HTML]{20124D} hamburger}    & {\color[HTML]{20124D} 1} & {\color[HTML]{20124D} 1} \\ \hline
{\color[HTML]{20124D} late}         & {\color[HTML]{20124D} 0} & {\color[HTML]{20124D} 1} \\ \hline
{\color[HTML]{20124D} service}      & {\color[HTML]{20124D} 0} & {\color[HTML]{20124D} 1} \\ \hline
{\color[HTML]{20124D} slow}         & {\color[HTML]{20124D} 0} & {\color[HTML]{20124D} 1} \\ \hline
\end{tabular}
\end{center}
\end{table}
Vector form of Tweet 1 =  [1, 1, 1,  1, 1, 1, 1, 1, 0, 0, 0] and 
Tweet 2 = [1, 0, 0, 0, 1, 0, 0, 1, 1, 1, 1].\\
\end{itemize}

\begin{itemize}
\item \textit{Term Frequency - Inverse Document Frequency (TF-IDF):} It is used to find the important terms or words that appear in the document or tweet based on their frequency. In TF-IDF, the less frequent word means more important. The formula of TF-IDF is TF multiplied by IDF.
\[ (tf-idf)_{t,d} = tf_{t,d} * {idf_t}\]
Term Frequency (TF) returns the frequency of a term (t) in each document (d) from the pre-processed vocabulary.\[ tf_{t,d} =\frac{n_{t,d}}{\sum_{k} n_{t,d}} \]
There, n = Number of times the term (t) found in the document (d). \\$\sum_{k} n_{t,d} = $ Total number of terms (t) in the document (d).
\begin{table}[hbt!]
\caption{Term Frequency for Tweet 1 and Tweet 2}
\label{tab:my-table}
\begin{center}

\begin{tabular}{|l|c|c|c|c|}
\hline
\textit{} &
  \multicolumn{2}{c|}{\textit{\begin{tabular}[c]{@{}c@{}}Tweet 1\\ $\sum_{k} n_{t,d} = 8 $\end{tabular}}} &
  \multicolumn{2}{c|}{\textit{\begin{tabular}[c]{@{}c@{}}Tweet 2\\ $\sum_{k} n_{t,d} = 6$\end{tabular}}} \\ \hline
\textbf{Terms} & \textbf{$n_{t,d}$} & \textbf{$tf_{t,d}$} & \textbf{$n_{t,d}$} & \textbf{$tf_{t,d}$} \\ \hline
delicious      & 1             & 1/8         & 1             & 1/6         \\ \hline
beef           & 1             & 1/8         & 0             & 0           \\ \hline
cheese         & 1             & 1/8         & 0             & 0           \\ \hline
burger         & 1             & 1/8         & 0             & 0           \\ \hline
mcdonald       & 1             & 1/8         & 1             & 1/6         \\ \hline
taste          & 1             & 1/8         & 0             & 0           \\ \hline
cheeseburger   & 1             & 1/8         & 0             & 0           \\ \hline
hamburger      & 1             & 1/8         & 1             & 1/6         \\ \hline
late           & 0             & 0           & 1             & 1/6         \\ \hline
service        & 0             & 0           & 1             & 1/6         \\ \hline
slow           & 0             & 0           & 1             & 1/6         \\ \hline
\end{tabular}
\end{center}
\end{table}\\
Inverse Document Frequency (IDF) calculates the weight of important words that appear in all documents.  
\[idf_t = \log \frac{N}{df_t} \]There, N = Total number of documents,\\ df = Number of documents containing term (t).

\begin{table}[hbt!]
\caption{Inverse Document Frequency for Tweet 1 and Tweet 2}
\label{tab:my-table}
\begin{center}
\begin{tabular}{|l|c|c|}
\hline
\textit{}      & \multicolumn{2}{c|}{{\color[HTML]{20124D} \textit{\textbf{N = 2}}}}        \\ \hline
\textbf{Terms} & {\color[HTML]{20124D} \textbf{$df_t$}} & {\color[HTML]{20124D} \textbf{$idf_{t}$}} \\ \hline
delicious & {\color[HTML]{20124D} 2} & {\color[HTML]{20124D} 0}    \\ \hline
beef      & {\color[HTML]{20124D} 1} & {\color[HTML]{20124D} 0.69} \\ \hline
cheese    & {\color[HTML]{20124D} 1} & {\color[HTML]{20124D} 0.69} \\ \hline
burger    & {\color[HTML]{20124D} 1} & {\color[HTML]{20124D} 0.69} \\ \hline
mcdonald  & {\color[HTML]{20124D} 2} & {\color[HTML]{20124D} 0}    \\ \hline
taste     & {\color[HTML]{20124D} 1} & {\color[HTML]{20124D} 0.69} \\ \hline
cheeseburger   & {\color[HTML]{20124D} 1}             & {\color[HTML]{20124D} 0.69}         \\ \hline
hamburger & {\color[HTML]{20124D} 2} & {\color[HTML]{20124D} 0}    \\ \hline
late      & {\color[HTML]{20124D} 1} & {\color[HTML]{20124D} 0.69} \\ \hline
service   & {\color[HTML]{20124D} 1} & {\color[HTML]{20124D} 0.69} \\ \hline
slow      & {\color[HTML]{20124D} 1} & {\color[HTML]{20124D} 0.69} \\ \hline
\end{tabular} \end{center}
\end{table}

% TABLE V  ----->  TERM FREQUENCY- INVERSE DOCUMENT FREQUENCY FORTWEET1ANDTWEET2
\begin{table}[hbt!]
\caption{Term frequency - Inverse document frequency for Tweet 1 and Tweet 2}
\label{tab:my-table}
\begin{tabular}{|l|c|c|c|c|c|}
\hline
\multicolumn{1}{|c|}{\textit{\textbf{}}} &
  \multicolumn{2}{c|}{\textit{$tf_{t,d}$}} &
  \textbf{} &
  \multicolumn{2}{c|}{{\color[HTML]{20124D} \textit{$(tf-idf)_{t,d}$}}} \\ \hline
\textbf{Terms} &
  \textbf{Tweet 1} &
  \textbf{Tweet 2} &
  \textbf{$idf_t$} &
  {\color[HTML]{20124D} \textbf{Tweet 1}} &
  {\color[HTML]{20124D} \textbf{Tweet 2}} \\ \hline
delicious &
  {\color[HTML]{20124D} 1/8} &
  {\color[HTML]{20124D} 1/6} &
  {\color[HTML]{20124D} 0} &
  {\color[HTML]{20124D} 0} &
  {\color[HTML]{20124D} 0} \\ \hline
beef &
  {\color[HTML]{20124D} 1/8} &
  {\color[HTML]{20124D} 0} &
  {\color[HTML]{20124D} 0.69} &
  {\color[HTML]{20124D} 0.0863} &
  {\color[HTML]{20124D} 0} \\ \hline
cheese &
  {\color[HTML]{20124D} 1/8} &
  {\color[HTML]{20124D} 0} &
  {\color[HTML]{20124D} 0.69} &
  {\color[HTML]{20124D} 0.0863} &
  {\color[HTML]{20124D} 0} \\ \hline
burger &
  {\color[HTML]{20124D} 1/8} &
  {\color[HTML]{20124D} 0} &
  {\color[HTML]{20124D} 0.69} &
  {\color[HTML]{20124D} 0.0863} &
  {\color[HTML]{20124D} 0} \\ \hline
mcdonald &
  {\color[HTML]{20124D} 1/8} &
  {\color[HTML]{20124D} 1/6} &
  {\color[HTML]{20124D} 0} &
  {\color[HTML]{20124D} 0} &
  {\color[HTML]{20124D} 0} \\ \hline
taste &
  {\color[HTML]{20124D} 1/8} &
  {\color[HTML]{20124D} 0} &
  {\color[HTML]{20124D} 0.69} &
  {\color[HTML]{20124D} 0.0863} &
  {\color[HTML]{20124D} 0} \\ \hline
cheeseburger & {\color[HTML]{20124D} 1/8} & {\color[HTML]{20124D} 0} & {\color[HTML]{20124D} 0.69} & {\color[HTML]{20124D} 0.0863} & {\color[HTML]{20124D} 0} \\ \hline
hamburger &
  {\color[HTML]{20124D} 1/8} &
  {\color[HTML]{20124D} 1/6} &
  {\color[HTML]{20124D} 0} &
  {\color[HTML]{20124D} 0} &
  {\color[HTML]{20124D} 0} \\ \hline
late &
  {\color[HTML]{20124D} 0} &
  {\color[HTML]{20124D} 1/6} &
  {\color[HTML]{20124D} 0.69} &
  {\color[HTML]{20124D} 0} &
  {\color[HTML]{20124D} 0.115} \\ \hline
service &
  {\color[HTML]{20124D} 0} &
  {\color[HTML]{20124D} 1/6} &
  {\color[HTML]{20124D} 0.69} &
  {\color[HTML]{20124D} 0} &
  {\color[HTML]{20124D} 0.115} \\ \hline
slow &
  {\color[HTML]{20124D} 0} &
  {\color[HTML]{20124D} 1/6} &
  {\color[HTML]{20124D} 0.69} &
  {\color[HTML]{20124D} 0} &
  {\color[HTML]{20124D} 0.115} \\ \hline
\end{tabular} 
\end{table}

%END TABLE ----->  TERM FREQUENCY- INVERSE DOCUMENT FREQUENCY FORTWEET1ANDTWEET2
In the end, Tweet 1 = [0, 0.863, 0.863, 0.863, 0, 0.863, 0.863, 0, 0, 0, 0] and Tweet 2 = [0, 0, 0, 0, 0, 0, 0, 0, 0.115, 0.115, 0.115]. Table V shows the final calculation of TF-IDF.
\end{itemize}

\subsection{Classification}
We have used the Train-Test-Split technique to divide the dataset into 75\% for training and 25\% for testing. Then applied different classification algorithms of Supervised Machine Learning on training data to train Machine Learning Classifiers and tested with testing data. Applied algorithms are Support Vector Machine, Multinomial Naive Bayes, Random Forest, and Logistic Regression.

\section{Result and Discussion}
We have evaluated the performance of our approaches with Accuracy, Precision, Recall, and F1-Score matrices. As the dataset was an imbalanced dataset, we have calculated the weighted average of precision, recall, and F1-Score. Formulas used for evaluation are as follows.

$ Precision = \frac{TP}{TP+ FP}  $\\ 

$Recall = \frac{TP}{TP+ FN} $ \\ 

$F1-Score  = \frac{2*Precision*Recall}{Precision+Recall}$ \\ 

$ Accuracy = \frac{Number\ of\ Correctly\ Predicted\ Data}{Total\ Number\ of\ Data} $\\

Table VI and VII show the summary of Accuracy, Precision, Recall, and F1-Score matrices found from applied Machine Learning classification algorithms and NLP techniques. Where Both SVM and Logistic Regression provide the highest accuracy of 77\% with a slight difference in F1-Score. 
%Table 7 Start
\begin{table}[hbt!]
        \caption{Classification algorithms with Bag-of-Words (BoW)}
        \label{tab:my-table}
    \begin{tabular}{|l|c|c|c|c|}
        \hline
        \multicolumn{1}{|c|}{\textbf{Algorithm}} &
        \textbf{\begin{tabular}[c]{@{}c@{}}Accuracy \end{tabular}} &
        \textbf{\begin{tabular}[c]{@{}c@{}}Precision \end{tabular}} &
          \textbf{\begin{tabular}[c]{@{}c@{}}Recall\end{tabular}} &
        \textbf{\begin{tabular}[c]{@{}c@{}}F1-Score\end{tabular}} \\ \hline
            \begin{tabular}[c]{@{}l@{}}Support Vector \\ Machine (SVM)\end{tabular}  & \textbf{0.77} & 0.76 & 0.77 & 0.75 \\ \hline
            \begin{tabular}[c]{@{}l@{}}Multinomial Naive\\  Bayes\end{tabular} & 0.74          & 0.72 & 0.74 & 0.72 \\ \hline
            \begin{tabular}[c]{@{}l@{}}Random \\Forest\end{tabular}          & 0.74          & 0.73 & 0.74 & 0.73 \\ \hline
            \begin{tabular}[c]{@{}l@{}}Logistic\\ Regression\end{tabular}    & \textbf{0.77} & 0.77 & 0.77 & 0.77 \\ \hline
    \end{tabular}
\end{table}
%Table 7 end

%Table 9 Start
\begin{table}[hbt!]
\caption{Classification algorithms with Term Frequency - Inverse Document Frequency (TF-IDF)}
\label{tab:my-table}
    \begin{tabular}{|l|c|c|c|c|}
        \hline
        \multicolumn{1}{|c|}{\textbf{Algorithm}} &
         \textbf{\begin{tabular}[c]{@{}c@{}}Accuracy\end{tabular}} &
         \textbf{\begin{tabular}[c]{@{}c@{}}Precision\end{tabular}} &
         \textbf{\begin{tabular}[c]{@{}c@{}}Recall\end{tabular}} &
         \textbf{\begin{tabular}[c]{@{}c@{}}F1-Score\end{tabular}} \\ \hline
            \begin{tabular}[c]{@{}l@{}}Support Vector\\  Machine (SVM)\end{tabular}  & \textbf{0.77} & 0.76 & 0.77 & 0.74 \\ \hline
             \begin{tabular}[c]{@{}l@{}}Multinomial Naive \\ Bayes\end{tabular} & 0.70          & 0.72 & 0.70 & 0.63 \\ \hline
            \begin{tabular}[c]{@{}l@{}}Random\\ Forest\end{tabular}          & 0.75          & 0.73 & 0.75 & 0.73 \\ \hline
            \begin{tabular}[c]{@{}l@{}}Logistic\\ Regression\end{tabular}    & \textbf{0.77} & 0.76 & 0.77 & 0.76 \\ \hline
    \end{tabular}
\end{table}
%Table 9 END
 Finally, In Table VIII we have compared the Accuracy and F1-Score between classification algorithms with BoW and classification algorithms with TF-IDF and selected the best approaches for Sentiment Analysis based on our experiments. In our approaches, the SVM and Logistic Regression provide the highest accuracy of 77\% with the Bag-of-Words technique.

%  Comparison Between Result Tables
 \begin{table}[hbt!]
\caption{Comparison Between Approaches of Table VI and Table VII}
\label{tab:my-table}
\begin{tabular}{|l|l|l|l|l|}
\hline
\textbf{}                                                              & \multicolumn{2}{l|}{Performance with BoW} & \multicolumn{2}{l|}{Performance with TF-IDF} \\ \hline
\textbf{Algorithms}                                           & \textbf{Accuracy} & \textbf{F1-Score} & \textbf{Accuracy} & \textbf{F1-Score} \\ \hline
\begin{tabular}[c]{@{}l@{}}Support Vector\\ Machine (SVM)\end{tabular} & 0.77                & \textbf{0.75}                & 0.77                  & 0.74                 \\ \hline
\begin{tabular}[c]{@{}l@{}}Logistic\\ Regression\end{tabular} & 0.77              & \textbf{0.77}     & 0.77              & 0.76              \\ \hline
\end{tabular}
\end{table}
% In Table VI and VII, Both SVM and Logistic Regression provide the highest performance of 77\% accuracy  with a slight difference in F1-Score. In  our approaches both Support vector machine and Logistic regression with Bag-of-Words provide the highest accuracy of 77\% with 75\% and 77\% of F1-Score respectively.

%Comparative Study Table%

\begin{itemize}
\item \textit{Comparison Between Related Works and Approaches of This Paper:
}% \begin{center}
\begin{table}[hbt!]
\caption{COMPARATIVE STUDY WITH RELATED WORK
}
\label{tab:my-table}
\begin{tabular}{|l|r|}
\hline
\multicolumn{1}{|c|}{\textbf{Title \& Year}} &
  \multicolumn{1}{c|}{\textbf{Algorithm \&  Accuracy}} \\ \hline
\begin{tabular}[c]{@{}l@{}}Sentiment Analysis Using Naive \\Bayes Algorithm Of The Data \\Crawler: Twitter (2019) {[}5{]}\end{tabular} &
  Support Vector Machine 63.99\% \\ \hline
\begin{tabular}[c]{@{}l@{}}Twitter Sentiments Analysis Using \\ Machine Learning Methods (2020) \\{[}6{]}\end{tabular} &
  Support Vector Machine 74.60\% \\ \hline
\begin{tabular}[c]{@{}l@{}}Sentiment Analysis for Airline \\Tweets Utilizing Machine Learning \\Techniques (2021) {[}7{]}\end{tabular} &
  Support Vector Machine 74.24\% \\ \hline
\begin{tabular}[c]{@{}l@{}}An Efficient Approach for Sentiment \\ Analysis Using Machine Learning \\ Algorithm (2020) {[}8{]}\end{tabular} &
  Support Vector Machine 68.00\% \\ \hline
\begin{tabular}[c]{@{}l@{}}Collaborative Classification Appr-\\oach for Airline Tweets Using \\Sentiment Analysis (2021) \\{[}9{]}\end{tabular} &
  \begin{tabular}[c]{@{}r@{}}Support Vector Machine 65.59\%\\ Logistic Regression 77.42\%\\ Random Forest 75.29\%\end{tabular} \\ \hline
\begin{tabular}[c]{@{}l@{}}A Comparative Analysis of Various \\ Machine Learning Based Social Me-\\dia Sentiment Analysis and Opinion \\Mining Approaches (2020) {[}10{]}\end{tabular} &
  \begin{tabular}[c]{@{}r@{}}Support Vector Machine 50.00\%\\ Logistic Regression 74.10\%\\ Random Forest 70.90\%\end{tabular} \\ \hline
\begin{tabular}[c]{@{}l@{}}A Study on The Performance of \\ Supervised Algorithms for Classifi-\\cation in Sentiment Analysis (2019)\\ {[}11{]}\end{tabular} &
  \begin{tabular}[c]{@{}r@{}}Support Vector Machine 66.59\%\\ Random Forest 49.67\%\end{tabular} \\ \hline
\begin{tabular}[c]{@{}l@{}}Sentiment Analysis of Arabic and\\  English Tweets(2019)\\ {[}12{]}\end{tabular} &
  \begin{tabular}[c]{@{}r@{}}Multinomial Naive Bayes 70.00\%\\ Logistic Regression 74.00\%\end{tabular} \\ \hline
The approaches of this paper &
 \begin{tabular}[c]{@{}r@{}}Support Vector Machine 77.00\%\\ Logistic Regression 77.00\%\end{tabular} \\ \hline
\end{tabular}
\end{table}

%Table IX Discussion\\
In Table IX, we have compared the accuracy of our selected approaches with some recent related work. In [5] and [8], Authors applied different data pre-processing techniques and ML algorithms. In our approach SVM provides 13\% and 9\% more accuracy respectively. In [9], Authors applied different algorithms and proposed a voting classifier. In our paper we have proposed more accurate approaches. Succinctly From [5] to [12] different authors have applied various techniques and different ML algorithms. But, the mentioned approaches comparatively provide better performance than existing studies.
\end{itemize}

\section{Conclusion}
 In this paper, we have implemented various Machine Learning classification algorithms and NLP techniques on a large, imbalanced, multi-classed, and real-world dataset to analyze sentiment. Our best approaches provide 77\% accuracy with both Support Vector Machine and Logistic Regression algorithm along with the Bag-of-Words technique. In the future, we would like to apply more advanced techniques to increase accuracy and will also try to build a generalized and robust model for similar datasets.

\section*{}

\vspace{12pt}

\end{document}